\documentclass[runningheads]{llncs} 
\usepackage{graphicx}
\usepackage{cite}

\begin{document}
\title{Deep Neural Ranking for Crowdsourced Geopolitical Event Forecasting}
\author{Giuseppe Nebbione\inst{1}\and
Derek Doran\inst{2} \and
Srikanth Nadella\inst{3} \and
Brandon Minnery\inst{3} }
\authorrunning{G. Nebbione et al.}
\institute{Dept. of Electrical \& Computer Engineering, University of Pavia, Italy \and
Dept. of Computer Science \& Engineering, \\ Wright State University, Dayton, OH, USA \and
Wright State Research Institute, Dayton, OH, USA
\email{giuseppe.nebbione01@universitadipavia.it \\ \{derek.doran, srikanth.nadella, brandon.minnery\}@wright.edu}}
\maketitle              % typeset the header of the contribution
\begin{abstract}
There are many examples of ``wisdom of the crowd" effects in which the large number of participants imparts 
confidence in the collective judgment of the crowd. But how do we form 
an aggregated judgment when the size of the crowd is limited? 
Whose judgments do we include, and whose do we accord the most weight? 
This paper considers this problem in the context of geopolitical
event forecasting, where volunteer analysts are queried to
give their expertise, confidence, and predictions about the outcome of
an event. We develop a forecast aggregation model that integrates topical 
information about a question, meta-data about a pair of forecasters, and their predictions
in a deep siamese neural network that decides which forecasters' predictions are more
likely to be close to the correct response. A ranking of the forecasters
is induced from a tournament of pair-wise forecaster comparisons, with the ranking used to create
an aggregate forecast. Preliminary results find the aggregate prediction of the best
forecasters ranked by our deep siamese network model consistently beats typical 
aggregation techniques by Brier score. 
\end{abstract}
\section{Introduction}
The science (and art) of forecasting has been studied in domains such as computer and mobile network monitoring and evaluation~\cite{f14,ren15,chen10},
meteorology~\cite{brier1950verification}, 
economics~\cite{bulligan2015forecasting}, sports~\cite{leitner2010forecasting}, 
finance~\cite{brock2018causality}, and geopolitics~\cite{collins1995prediction, cahnman1942methods}.
A closely related line of research involves leveraging the power of aggregation to improve forecast accuracy. In this approach, multiple forecasts 
from different sources (human or algorithm) are merged to achieve results that are, on average, superior 
to those of even the best individual forecaster.  This ``wisdom of crowds" effect~\cite{surowiecki2005wisdom} 
arises from the fact that individual forecasters possess different information and biases~\cite{makridakis1983averages}, leading 
to uncorrelated (or better yet, negatively correlated) prediction errors that cancel when combined (e.g., by averaging).  
Implicit in the above description is the assumption that forecasters are more than just proverbial ``dart-throwing chimps": 
they possess some degree of relevant knowledge and expertise enabling them to perform above chance. By extension, not all forecasters 
have equal expertise, suggesting that it may be possible to identify smaller, ``wiser" ensembles of 
the most skilled forecasters~\cite{goldstein14} \textemdash or alternatively, to assign different weights to different forecasters 
based on their expected contribution to the crowd's accuracy.

Indeed, a common approach to forecast aggregation is to use weighted
models that favor the opinion of forecasters by their experience and past accuracy. 
Yet even weight aggregation methods are imperfect. This is because they
tend to oversimplify the forecaster representation and do not take into
account how skilled a forecaster may be within a specific context.
This issue becomes even more apparent when considering the problem of 
{\em geopolitical forecasting}~\cite{warnaar2012aggregative},
where an analyst is asked to predict 
the outcome of an international social or political event. In geopolitics, a 
substantial number of factors, some of which may be highly unpredictable, 
must be considered. 
While a forecaster may possess expertise about a particular international region, political
regime, or event type
(e.g., Russian military operations in Ukraine), 
the countless factors and hidden information contributing
to an event outcome guarantee that expertise will correlate imperfectly with prediction accuracy. 
Moreover, such expertise may not generalize to other forecasting questions involving
different regions, regimes, or event types. We thus suggest a deep learning solution to the problem of 
crowdsourced geopolitical forecaster aggregation. 
By the architecture of its network layers, and the number
and size of such layers, deep neural networks can be designed 
to become arbitrarily expressive~\cite{lecun2015deep}. Such expressiveness 
is necessary in identifying good predictions from a crowd of
forecasters because the factors that determine the accuracy of a prediction are numerous and latent.
A challenge of applying expressive deep networks, however, is the 
expense of acquiring the substantial amount of data needed
to train a generalizable model. The data that do exist come from
past forecasting competitions~\cite{grela2015crowdsourcing}, yet such data may still 
not be adequate to train a deep network with
sufficient expressive power. 

To overcome this challenge, we propose a 
method for geopolitical prediction aggregation
from a crowd based on a {\em neural ranking} of forecasters.
This refers to a deep neural network that
defines a ranking of forecasters based on their {\em predicted relative accuracy}
for a given forecasting question. Forecasters' rankings are based on their performance in previous 
questions, their self-reported confidence in their prediction, and 
on the latent topics present in the 
previous questions they performed well on. More formally,  
given a set of $m$ geopolitical questions asked to $n$ forecasters, where some subset of 
$k < n$ forecasters provide a prediction to each question, a neural ranker
yields a {\em ranking} of the $n$ forecasters based on how
likely their response is to be ``closest" to the correct outcome. The ranking
is derived from a tournament where all pairs of ${k \choose 2}$ forecasters
are compared by a deep siamese network to identify the one whose forecast
is more likely to be correct. This scheme also yields a training dataset of 
$\sum_{i=1}^m {k_i \choose 2}$ pairs of examples for training and testing the
deep siamese network, which is far more likely to be of sufficient size
for training a generalizable model. We evaluate our 
neural ranking approach against a standard weighted aggregation algorithm in which weights are based on forecasters'
average past performance.
Using data provided by the IARPA Good Judgement Project~\cite{atanasov2016distilling}, we find
that our neural ranking method produces substantially improved Brier scores (a measure of prediction accuracy) 
compared to the standard weighted aggregation model. 

The layout of this paper is as follows: Section~\ref{sec:rr} 
gives details about the state of the art in the context of aggregation methods 
for forecasts. Section~\ref{sec:meth} details the design of our neural 
ranking system and its siamese network and ranking components. 
Section~\ref{sec:eval} describes evaluation results for our system compared to standard weighted aggregation
models. Section~\ref{sec:Conclusion} concludes the paper and offers directions for future research.

\section{Related Work} \label{sec:rr}
Sir Francis Galton published the first modern scientific study of crowd wisdom more than a century ago in a paper that 
analyzed data from a contest in which hundreds of individuals independently attempted to estimate the weight of a 
prize-winning ox~\cite{galton1907vox}. More recent research has similarly capitalized on data from large-scale public 
or private forecasting contests, where forecasters compete (often for financial award) individually or in teams to produce the most accurate predictions
within a particular domain. Popular examples of such forecasting tournaments include fantasy sports~\cite{bhatt2017enhancing}
~\cite{goldstein14} and, more recently, geopolitics~\cite{tetlock2014forecasting}. These tournaments, which can 
include thousands of competitors, are a rich source of quality data because they 
naturally attract participants who are both knowledgeable and motivated.

With respect to forecast aggregation, recent work has focused on integrating forecaster features and past performance. 
Budescu {\em et al.}~\cite{budescu2014identifying} proposes a method for measuring
a single judge's (e.g., a single forecaster's) contribution 
to the crowd's performance and uses positive contributors 
to build a weighting model for aggregating forecasts. 
Forlines {\em et al.}~\cite{forlines2012heuristics} shows how 
heuristic rules based on the qualities of each forecaster 
can significantly improve aggregated predictions. 
Hosen {\em et al.}\cite{hosen2015improving} applies a neural network to forecast 
a range of likely outcomes based on a weighted average aggregation 
of forecaster predictions. Further work by 
Ramos~\cite{ramos2014robust} uses a particular
nonparametric density estimation technique called L2E, which aims at making the
aggregation robust to clusters of opinions and dramatic changes. 
The Good Judgement project by Tetlock {\em et al.}~\cite{tetlock2014forecasting} introduces
an approach based on a novel cognitive-debiasing training design. 
Atanasov {\em et al.}~\cite{atanasov2016distilling}
shows that team prediction polls outperform prediction markets once
forecasts are statistically aggregated using various techniques such as temporal 
decay, differential weighting based on past performance, and recalibration. 

In this work, we consider an innovative method to tap into the wisdom of the crowd
for geopolitical forecasting. We reformulate the
crowd forecasting problem as a neural ranking problem, in which a deep neural network is used to rank forecasters
based on their expected
relative accuracy for a given forecasting question. The neural network learns a complex representation of 
forecasters, their forecasts, and contextual information about the question of interest. The resulting ranking is
then used to create a weighted aggregation of forecasts (i.e., a crowd forecast) for each unique forecasting question.

\section{Methodology} \label{sec:meth}
\begin{figure}
\includegraphics[width=\textwidth]{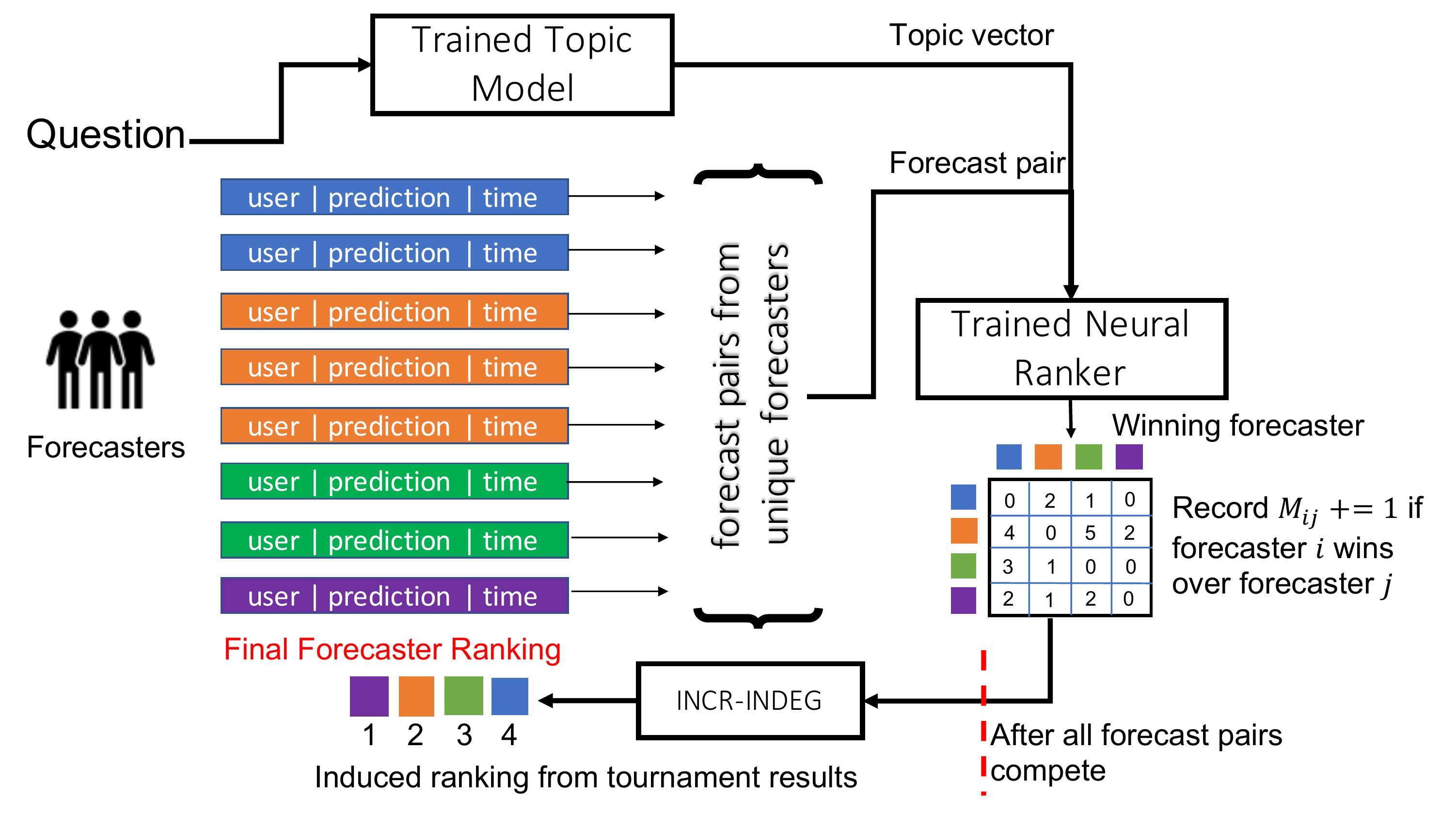}
\caption{Approach overview. Given multiple
predictions on the outcome of a question from multiple forecasters, 
we apply a siamese neural network to identify, for all
pairs of predictions submitted by unique forecasters, which of the two is more likely to be 
closer to the true outcome. The network additionally considers the latent topics within the question text
to establish relationships between forecasters and the topics they make predictions over. 
Comparisons are recorded in a matrix that can be represented 
as a weighted tournament over the forecasters. The INCR-INDEG algorithm is applied to the
tournament to infer a ranking. The predictions made by those forecasters in a top 
percentile of the ranking are aggregated to form the crowd's prediction.} 
\label{fig:overview}
\end{figure}
We introduce a new crowd aggregation technique based on a novel 
{\em ranking} of forecasters who have provided
a series of predictions for a particular question as illustrated in Figure~\ref{fig:overview}. We assume that forecasters can 
submit multiple predictions for a given question; i.e., forecasters are allowed to update their predictions as they receive new information over time.  
All pairs of forecasts made by unique forecasters are passed into a 
deep siamese network~\cite{koch2015siamese} that will
evaluate which of the two forecasts are most likely to be closer to the true event outcome. 
The submitted forecasts include a forecaster-provided confidence score
about their prediction. The siamese network further considers the kinds
of topics featured in each question, which is derived by running the 
question text through a trained topic model. The outcome of all comparisons
by the neural ranker is composed in a matrix $M$ that defines a tournament 
graph over the forecasters. The tournament graph is processed through the INCR-INDEG
algorithm~\cite{coppersmith2006ordering} to induce a ranking of forecasters by their comparative ability
to provide better predictions. An aggregation of the highest ranking forecasters
(defined by some top percentile of the ranking) are used as the crowd's prediction. In the event
that multiple forecasts are submitted by the same ranker, we compute a prediction conditioned on
a time $t$ by using the latest prediction submitted by each forecaster prior to $t$. We elaborate on 
the topic model, neural ranker, and ranking algorithm next. 

\subsection{Topic Modeling}
The first step is to learn a topic model over a corpus of questions about geopolitical events that have been asked to a 
to a crowd of forecasters. The purpose of the topic model is to identify {\em a priori} the 
topics latent within the set of questions asked to forecasters. In doing so, we seek
to provide the neural ranker with {\em topical} information about the question being asked,
so that the ranker becomes able to learn associations between forecasters and the 
topics that they are (not) proficient in. We specifically apply Latent Dirichlet Allocation (LDA)~\cite{blei2003latent} across an 
entire corpus of geopolitical questions that have been asked to forecasters in the past. 
LDA is a widely used unsupervised learning technique that represents 
a topic $t_j$  ($1 \le j \le T$) as a multinomial probability distribution
$p(w_i|t_j )$ over $M$ words taken from a corpus $D = \{d_1,d_2,\cdots,d_N\}$ of documents
$d_i$ where words are drawn from a vocabulary $W = \{w_1, w_2,\cdots, w_M \}$.
The probability of observing a word in document is defined as
$p(w_i|d)=\sum_{j=1}^{T}p(w_i|t_j)p(t_j|d)$. Gibbs sampling is used to estimate the word-topic
distribution $p(w|t)$ and the topic-document distribution $p(t|d)$.
While various metrics to identify the ideal number of topics $T$ for a given 
corpus have been identified~\cite{griffiths2004finding,bolelli2009topic}, no metric is a silver bullet, and manual inspection 
is often the preferred approach. Manual inspection ensures that words clustered into various topics are collectively meaningful, 
in the sense that a reasonable person may look at the
words in a cluster to conclude that they are representative of a topic. For example, a cluster
of words including {\em nuclear, peninsula, dictatorship, DMZ} likely represents a topic 
about North Korea. Admitting a larger number of topics in the model carries the risk of producing 
clusters of words that are not topically related. A learned model over $T$ topics is then used to produce 
a $T$-dimensional topic vector for each question. The $i^{th}$ 
component of this vector simply represents the proportion of words in a question from topic $i$ of the model.

\subsection{Neural Ranker}
\begin{figure}
\includegraphics[width=\textwidth]{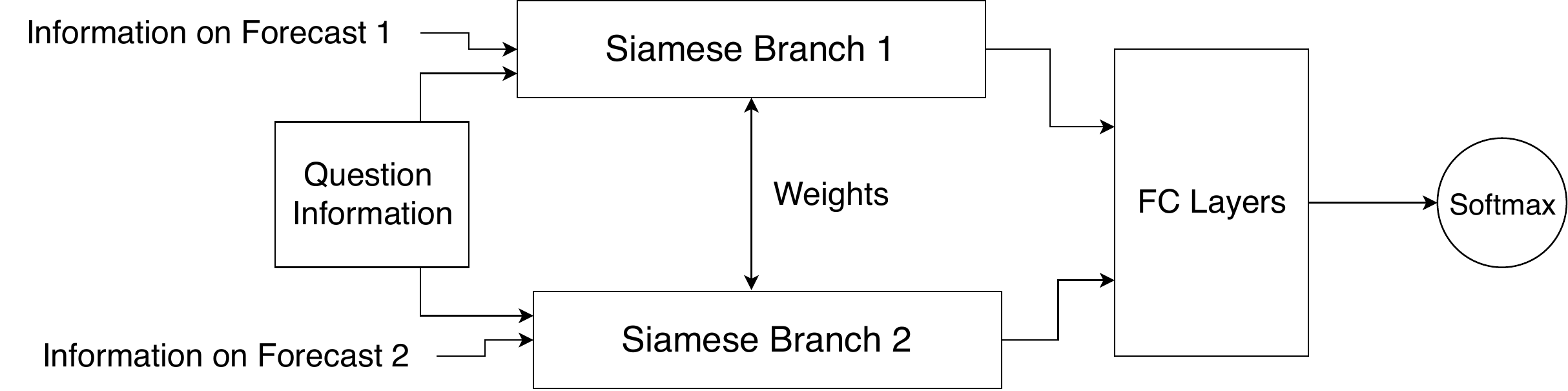}
\caption{Neural ranking architecture with two siamese branches. Both branches take 
the same topic vector of a question and share network weights. The output of 
a siamese branch can be thought of as a vector embedding of data about a forecaster, 
a question, and the forecaster's prediction for the question. A softmax output is interpreted as
how likely the first forecast is closer to the true answer than the second.} \label{fig:network}
\end{figure}

A pair of forecasts made and topic vector for a question are then fed into a neural ranker. 
A neural ranker that simply takes as input all forecasts and the question
to produce a ranking may be intuitively appealing, but may also 
not be possible in most geopolitical forecasting contexts. This is because
such a network would require a large amount of training data to fit its myriad of parameters while
ensuring generalization, and because the size of the input must be fixed. But for geopolitical 
forecasting,  it is difficult to find a dataset
consisting of thousands of questions with forecasts provided by a fixed 
set of forecasters; in fact there is often a variable number of forecasters 
who make predictions about a small number of questions~\cite{bolger2017use}. 

We instead consider a neural ranker that evaluates which of a pair of forecasts submitted for a question is more
likely to be closer to the correct response. By training such a ranker over forecasts submitted by the same set of 
forecasters across a set of questions with a variety of topics, associations between forecasters, question topics, 
and their prediction patterns can be learned and exploited for identifying the ``better'' of two predictions. 
The neural ranker is designed as a deep siamese network~\cite{hoffer2015deep,sun2014deep} illustrated in Figure~\ref{fig:network}. 
The siamese network is composed of two identical multi-layer perceptrons that share weights. 
Each branch takes as input a forecast concatenated with meta-data about the forecaster and the topic
vector quantifying the information content of each question. The output of a siamese branch is 
a vector embedding representative of information about the forecaster, the question asked, and their prediction. 
The embeddings from each branch are concatenated and fed into a series of fully connected layers. A 
softmax output scores whether the prediction of forecaster 1 is superior. The network is trained by 
stochastic gradient descent with momentum using a binary cross-entropy loss function. 

\subsection{Ranking process}
The trained neural ranker evaluates all pairs of predictions for a given question submitted by two unique forecasters.  For a question asked to $n$ forecasters, we define an $n \times n$ matrix $M$ where $M_{ij}$ counts the 
number of times forecaster $i$'s prediction was chosen over forecaster $j$'s. We then define the matrix
$T$ where $T_{ij} = M_{ij} / (M_{ij} + M_{ji})$ as the adjacency matrix of a weighted tournament with probability
constraints $T = (V, E, w)$. Here, $V$ represents forecasters and 
$w: E \rightarrow [0,1]$ is a weight function such that $w((v_i,v_j)) + w((v_j,v_i)) = 1$ 
for any $(v_i, v_j)\in E$. A ranking of the $V \in T$ can be defined by an ordering $\sigma : V \rightarrow {1, 2, ..., |V|}$ 
minimizing the sum of all $w((v_i, v_j))$ of all backedges induced from $\sigma$, where a backedge $(v_a,v_ b)$ has $\sigma(v_a) > \sigma(v_b)$. Finding $\sigma$ is NP-hard~\cite{alon06, charbit07}, but Coopersmith {\em et al.} 
discovered a simple 5-approximation algorithm called INCR-INDEG~\cite{coppersmith2006ordering} 
where $\sigma$ orders vertices by their weighted in-degree $D(v_i) = \sum_{v_j \in V \setminus {v_i}} w(v_i, v_j)$ with ties broken
randomly. We apply INCR-INDEG to $T$  to produce a final ranking of forecasters for a 
question. 

\section{Evaluation} \label{sec:eval}
We evaluate our neural ranker using public data from the IARPA Good Judgment Project\footnote{https://www.dni.gov/index.php/newsroom/press-releases/item/1751-iarpa-announces-publication-of-data-from-the-good-judgment-project}. This data is a product of a four-year long (2011-2015) prediction
tournament over geopolitical events. Questions from the project were 
published for a variable amount of time, during which a forecaster could
submit a prediction along with a confidence score (an integer value between 1 and 5). 
A forecaster could update his or her forecast at any time, and the most
recent forecast on any given day was taken as the forecast for that day.
For each question, forecasters were presented with disjoint possible
outcomes and were asked to submit the probability that each outcome would
occur. Forecasters were free to determine which questions to attempt and how often to update their forecasts,
resulting in a variable number of forecasters and forecasts per question.
Questions from each year of the project were answered by a different set
of forecasters, with a subset of forecasters participating across multiple years. 

% \begin{table}[h!]
%   \begin{center}
%     \caption{Dataset description}
%     \label{tab:table2}
%     \begin{tabular}{l|r|r|r|r} 
%       \textbf{Statistic} & \textbf{Year 1} & \textbf{Year 2} & \textbf{Year 3} & \textbf{Year 4}\\
%       \hline
%       number of questions & 101 & 172 & 233 & 158 \\
%       number of forecasters & 1,972 & 1,238 & 1,573 & 7,242 \\
%       number of forecasts & 146,220 & 181,884 & 302,392 & 572,322 \\
%     \end{tabular}
%   \end{center}
% \end{table}
Questions greatly varied in terms of subject but were always related
to determining if, when, or how a political, geographical, social, or economic event would occur in the future. Question also carried a brief but detailed description and links to news articles and on-line sources for a forecaster to begin investigations. Examples of questions from the IARPA forecasting tournament include: 
\begin{itemize}
\item {\em By 1 January 2012 will the Iraqi government sign a security agreement that allows US troops to remain in Iraq?} 
\begin{enumerate} 
\item Yes, by 15 October 2011
\item Yes, between 16 October and 1 January
\item No
\\
\end{enumerate}
\item {\em Will the United Nations Security Council pass a new resolution concerning Iran by 1 April 2012?}
\begin{enumerate}
\item Yes, a new resolution will be passed
\item No, a new resolution will not be passed
\\
\end{enumerate}
\item {\em Who will be inaugurated as President of Russia in 2012?} 
\begin{enumerate}
\item Medvedev
\item Putin
\item Neither
\end{enumerate}
\end{itemize}
The responses to each question are constructed so that the actual outcome will always 
correspond to exactly one of the responses. We analyzed our approach using questions from the first year of the competition 
(extensive evaluations across all four years will be pursued in future work). 
The 101 questions in the first year had an average of 1,440.52 forecasts with standard deviation of 668. 

%% Present the resulting topic model and representative keywords. Give an example of a question
%% where the dominant topic is of each type. This will provide some validity to the topic labels
%% that you manually derived. 
\subsection{Neural model evaluation}
We ran LDA to extract the latent topics from a corpus of 
each question's text concatenated with their description. Manual inspection of
words organized into clusters by LDA were used to determine that six topics would be appropriate. 
This was reached by a trial-and-error approach, starting from a large number
of topics, where we continued to reduce the number of topics in the model until 
there was insignificant overlap of words within different topics. 

The neural ranker is composed of two siamese branches of three fully connected layers. Each layer 
is composed of 32
ReLU activation functions. The 32 dimensional output of each branch is concatenated and
passed into four fully connected layers, each composed of 64 ReLU activation functions. 
The final fully connected layer outputs a single logistic activation. 
We carried a preliminary evaluation of the neural 
ranker using all 
questions from the first year of the
IARPA forecasting tournament featuring 101 questions or ``IFIPs".
The neural ranker was trained
over a random sampling of 5 million response pairs
across year 1\footnote{We must mention that the 5M response pairs were meant to be sampled from the first 50 questions of the
Year 1 data set, but due to a bug identified after submission of the paper, the sampling occurred across all Year 1 questions. 
This caused some testing set examples to have bled into our training data as well. However, this evaluation
bug does {\bf not} affect our evaluation of the crowd's performance -- which is the ultimate aim of the model -- to be discussed in 
Section 4.2. Our subsequent work will be carried out with this bug fixed.}. Model selection was performed by 
a validation set composed of all pairs of responses
from two questions (IFIP 1050 and 1051 in the
dataset, respectively). The network was trained with mini-batches 512 prediction pairs. Model parameters were optimized to minimize 
cross-entropy loss by stochastic gradient descent with a learning rate of $\lambda = 0.01$ and with a momentum 
term $\rho = 0.9$ added. The network was trained over multiple epochs and instantiated an early stoppage procedure when 
the validation error exceeded training error for more than 3 consecutive epochs. After training, we ultimately select the model parameter
settings from the end of the latest epoch whereby the training error rate did not fall below validation error
over IFIP 1050 and 1051 questions. 
The training process was repeated for a small number of different networks where the number of layers and size of the layers varied.
The layer sizes and counts listed above were selected based on these repeated experiments; a comprehensive 
sensitivity analysis of the rankings against different architectures will be the subject of future work. 

\begin{figure}
\includegraphics[width=\textwidth]{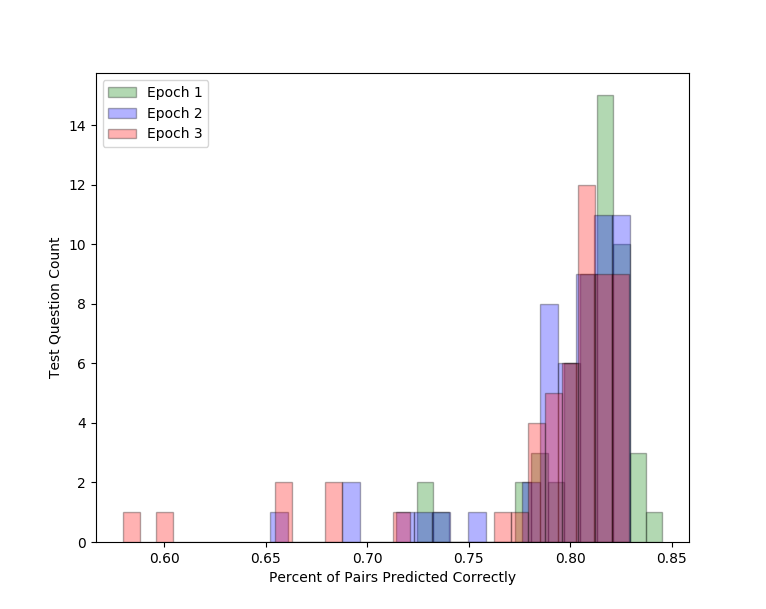}
\caption{Distribution of the proportion of forecast pairs where the neural 
ranker identifies the better prediction, per question in the validation
set. Colors correspond to the distribution after one, two, and three training
epochs. The distributions suggest decreasing validation set performance for later epochs, 
despite increasing accuracy over the training set.} 
\label{fig:nn_perf}
\end{figure}

The neural ranker was tested against all forecast pairs from the 54 questions asked in the Year 1 dataset,
constituting the later half of questions asked
to the forecast crowd. 
These questions were chosen to simulate a test set where questions asked in the future are evaluated on a model 
trained on examples from the past. Figure~\ref{fig:nn_perf} shows the distribution of the number of questions for which the neural ranker
is able to choose the better of two predictions
at some accuracy. The performance is encouraging: despite the variety and complexity of the questions asked, the neural ranker is able to identify 
the better of two predictions over 80\% of the time
on average. We found performance to decrease slightly after one epoch, while training accuracy 
continued to improve. We thus fixed the model after training for one epoch to minimize overfitting. 

\subsection{Crowd performance}
We defined a crowd's prediction as the weighted arithmetic mean prediction for each 
question, where weights were based on a forecaster's ranking for that 
question. For simplicity's sake, we used binary weights: that is, a forecaster $c_i$'s prediction was either included ($w_i=1$) or not ($w_i=0$) in the aggregate 
based on his or her rank computed by the neural ranker and INCR-INDEG algorithm.
We used a proper scoring rule called the Brier score~\cite{brier1950verification}, to measure the accuracy of a crowd’s prediction for a given question. The Brier score, or Mean Quadratic Score, measures forecasting accuracy under discrete choice conditions such as the answer sets for the questions used in the IARPA tournament. The Brier score is proper as it encourages forecasters to report their true beliefs (i.e., their best estimate of the likelihood of a future event occurring). It is calculated as: $$\frac{1}{n}\sum_{i=1}^n\sum_{j=1}^r (f_{ij} - o_{j})^2$$ where $n$ is the size of the crowd, $r$ is the number of possible results, $f_{ij}$ is the probability of result $j$ predicted by forecaster $i$ and $o_j \in \{0,1\}$ is equal to 1 only if result $j$ does occur. Brier scores can range from 0 to 2, with lower scores corresponding to greater accuracy. In the case of a question whose response options are ordered multinominals (e.g. ``less than 10"; ``between 10 and 20"; ``greater than 20"), an extension of the Brier score is required to ensure that more credit is awarded for choosing a response close to the true outcome. We use an adaptation of the Brier score by Jose {\em et al.}~\cite{jose2009sensitivity} for these questions.

Given ever-changing geopolitical landscapes, forecasters periodically and asynchronously update their predictions throughout the lifespan of a question. We thus calculate an aggregated prediction at the end of each day the question is open using the latest predictions from the top-ranked forecasters. A daily Brier score for each question is subsequently computed, and we define the ultimate performance of the crowd for a question as the mean of its daily Brier scores (MDB). The overall accuracy of an aggregation method over multiple questions is given by the mean of the MDBs, or the MMDB. 

\begin{figure}
\includegraphics[width=\textwidth]{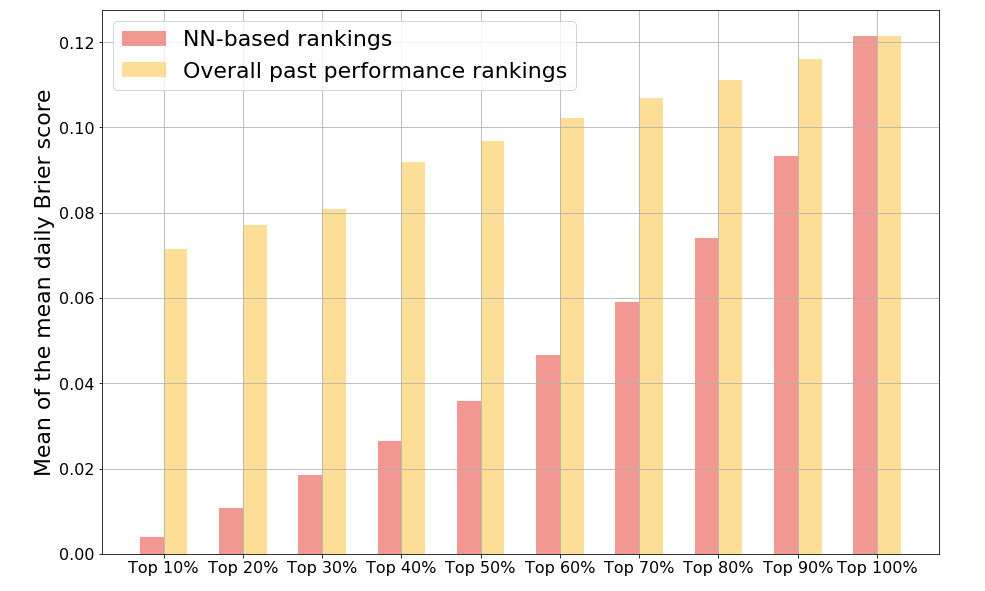}
\caption{Brier scores of crowd predictions based
on the neural ranker and by past performance.} 
\label{fig:mmdb}
\end{figure}

To explore how crowd accuracy varies as a function of the rank-based weights, we generated results using
 different ranking cutoffs, where a varying top percentage of ranked forecasters were assigned weight 
$w_i = 1$ with all others receiving weight 0. Figure~\ref{fig:mmdb} shows MMDB on the test set for different ranking cutoffs (orange bars). 
These results show that crowds composed of higher-ranked forecasters substantially outperformed crowds that 
included lower-ranked forecasters. However, it is reasonable to ask whether similar results could have been 
achieved using simpler methods. We thus compared our approach against a typical 
aggregation strategy (Figure~\ref{fig:mmdb}, yellow bars) in which forecaster weights $w_i$ are based on their MMDB scores 
computed at the end of each day while the question was
open for predictions by forecasters. 
This MMDB thus considers forecasts
for each day a question was open. 
As Figure~\ref{fig:mmdb} shows, the benchmark 
weighting scheme also yields more accurate forecasts compared to an  
unweighted aggregation (shown when using 100\% of the crowd); however, 
our NN-based weightings produced consistently superior results, and this differential increased
with greater selectivity.

\section{Conclusion}\label{sec:Conclusion}
This paper introduced a novel scheme to aggregate predictions from a crowd of
forecasters for predicting geopolitical events. Whereas the current art 
bases the relevance of predictions by forecasters' past performance, 
we score relevance by a question-specific ranking of forecasters 
induced by a tournament where forecaster predictions are 
adjudicated by a siamese neural network. Preliminary results show that
choosing the average prediction of forecasters in a top percentile 
of our neural ranking consistently yields a lower (superior) Brier score 
than using the same top percentile of forecasters based on past 
performance. 
% Part on Future Experiments, feel free to edit if something does not convince you
Future experiments may include attempting the ranking phase over the following years
and experimenting with different forecaster input representations, for example by including
psychometric features included in the Good Judgment data set.
Another interesting approach would be fine tuning the topic modeling for example by using more
robust approaches for the determination of the number of topics.

\section{Acknowledgements}
This work is supported in part by the Office of the Director of National Intelligence (ODNI), Intelligence Advanced Research Projects Activity (IARPA),via 2017‐17072100002 and by the University of Pavia through a mobility grant awarded to Giuseppe Nebbione.
The views and conclusions contained herein are those of the authors and should not be interpreted as necessarily representing the official policies, either expressed or implied, of ODNI, IARPA, or the U.S. Government. The U.S. Government is authorized to reproduce and distribute reprints for governmental purposes notwithstanding any copyright annotation therein.
\bibliographystyle{splncs04}
\bibliography{biblio}

\end{document}